\PassOptionsToPackage{square,sort,comma,numbers}{natbib}
\documentclass{article}
\usepackage[utf8]{inputenc}
\usepackage[final]{neurips_rl2019}
\usepackage[pdftex]{graphicx}
\graphicspath{{../pdf/}{../jpeg/}}
\usepackage[cmex10]{amsmath}

\usepackage{algorithm}
\usepackage{algpseudocode}
\usepackage{array}
\usepackage{mdwmath}
\usepackage{mdwtab}
\usepackage{subfig}
\usepackage{hyperref}

\title{Human-Robot Collaboration via Deep Reinforcement Learning of Real-World Interactions}
\date{September 2019}
\author{Jonas Tjomsland$^1$, Ali Shafti$^{1,2,3}$, A. Aldo Faisal$^{1,2,3,4}$\\ $^1$Dept. of Bioengineering, $^2$Dept. of Computing, $^3$Data Science Institute, \\$^4$UKRI CDT for AI in Healthcare, Imperial College London\\jt732@cam.ac.uk, a.shafti@imperial.ac.uk, aldo.faisal@imperial.ac.uk}

\begin{document}
\maketitle

\begin{abstract}
 \em{We present a robotic setup for real-world testing and evaluation of human-robot and human-human collaborative learning. Leveraging the sample-efficiency of the Soft Actor-Critic algorithm, we have implemented a robotic platform able to learn a non-trivial collaborative task with a human partner, without pre-training in simulation, and using only 30 minutes of real-world interactions. This enables us to study Human-Robot and Human-Human collaborative learning through real-world interactions. We present preliminary results, showing that state-of-the-art deep learning methods can take human-robot collaborative learning a step closer to that of humans interacting with each other.}
 
 
\end{abstract}

\section{Introduction}
\begin{figure}[h]
    \centering
    \includegraphics[width=7cm]{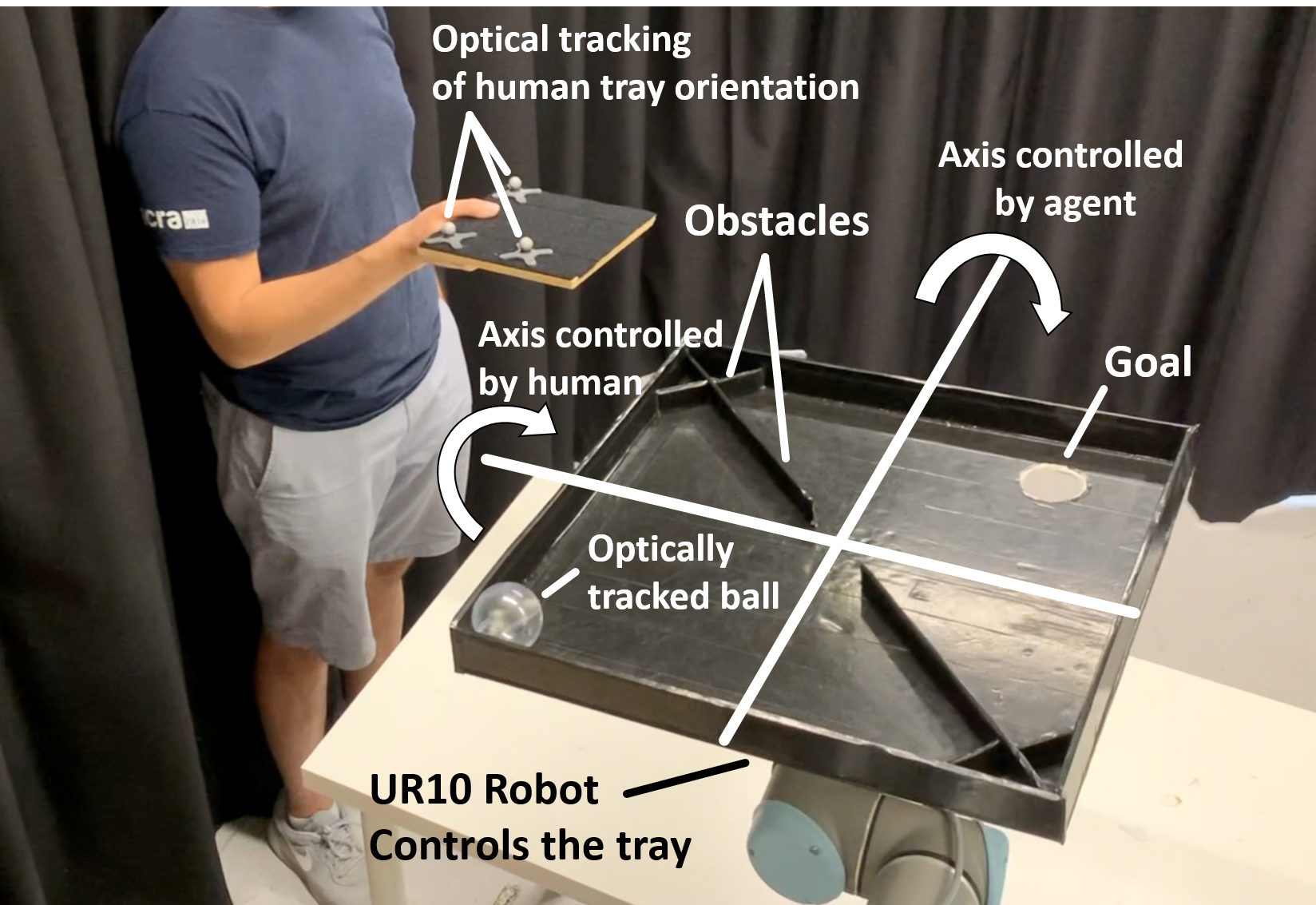}%
    \caption{The robotic setup used for human-robot and human-human collaboration.}%
    \label{robot}%
\end{figure}

Artificially intelligent agents are displaying impressive behaviour in diverse individual tasks, such as skin cancer classification \cite{deep} and complex board games \cite{AlphaGo}. Similarly, multi-agent environments, where a degree of teamwork is required, are being explored \cite{multiplayer}. Robots, on the other hand, have yet to achieve human level performance in a wide variety of real-world collaborative tasks. To approach this challenge, this work has created a robotic setup for controlled studies into human-robot collaborative learning. This is implemented as a deep reinforcement learning (DRL) agent in a non-trivial, physical human-robot collaboration task. See Fig. \ref{robot} for an illustration of the robotic setup including a human, a robot with a tray, and a DRL agent driving the robot.

For robots to achieve human level of performance in collaborative tasks, we need to include their human interaction partner, not only at deployment, but also during training. This way, robots will be able to build personalised models of human partners, a critical trait given the stochasticity of human behaviour. Moreover, human operators might have specific weaknesses that a robotic interaction partner should learn to compensate for. This sets multiple requirements to the simplicity and sample-efficiency of the training process.

Until recently, applying deep learning techniques with humans partaking in training has been impossible, due to the enormous amount of data needed to learn even the simplest task. Additionally, real-world training has been limited by the wear and tear of mechanical systems. This is why most deep learning algorithms are trained and evaluated in simulation. However, it is not straightforward to simulate the dynamics of a real world scenario involving a human collaborator and to effectively perform human-agent co-learning of real-world tasks within simulation. Taking learned behaviour from simulation to real-world is already a growing research field \cite{sim1, sim2, sim3}.

Within this paper we present a robotic setup enabling the human-robot team to not only solve a collaborative task within 30 minutes of real-world training, but to have a performance comparable to that of a human-human team for the same task. Ultimately showing that the sample-efficiency of state-of-the art DRL-methods now allows for human-in-the-loop training from scratch, opening the door to further studies on collaborative learning.

\section{Related Work}

The application of deep learning in human-robot interaction is not novel. Examples of this can be found in the field of shared autonomy; environments in which multiple agents (human or artificial) act at the same time, to achieve shared or individual goals. Chen et al. \cite{planning} developed a robot which was able to exhibit socially compliant behaviour using DRL. Reddy et al. \cite{shared}, also utilised DRL, to augment a human's actions to achieve better performance in a drone-flying task. They simulated general human models as pilots to handle the large amount of training samples required. Other works involving multi-agent systems has shown that multiple artificial agents can collaborate in complex computer games and outperform human teams \cite{openai}. All of the aforementioned approaches are either trained in simulation or operate solely in a simulated world.

Although today's most popular machine learning techniques require a large number of data, there are exceptions. PILCO \cite{Pilco}, is a model-based policy search method that leverages the sample efficiency of Gaussian processes to build dynamics models. With less than a minute of interaction time, PILCO can learn to stabilise classic control problems. However, significant computational power and time are needed to complete the policy search after interaction. In the case of DRL, the Soft Actor-Critic (SAC) algorithm with automatic temperature tuning is the state-of-the-art with respect to sample-efficiency. Haarnoja et al. \cite{SAC} illustrated this by showing how a four-legged robot could learn to walk from scratch, without pre-training in simulation, in under two hours. Crucial to this performance was the fact that no extensive hyperparameter tuning was needed, which drastically lowered the number of trials required.

\vspace{-3.5pt}
\section{Method}

\subsection{Robotic Setup}
The framework used to train and evaluate human-robot collaboration in this study, consists of three main components: A Linux workstation, a Universal Robots UR10 industrial robot \cite{UR10} with a tray attached and a motion capture system from Optitrack \cite{optitrack}. Running on the workstation, the Robot Operating System (ROS) is used to send and receive information between the components. A vital part of the setup is the \emph{jog\underline{ }arm} ROS package, developed by the University of Texas Nuclear and Applied Robotics group. The package simplifies the communication of smooth velocity commands to the UR10 robot. A tray with a rolling ball, some obstacles and a goal, is placed on top of the robot and establishes the arena for the interaction task. The task is solved purely by rotating the tray around the x-y axes; no translation is needed, and neither is rotation around the z-axis. See Fig. \ref{robot} (b) for a illustration of the tray's layout.

The tray is designed such that the task cannot be solved by controlling only one axis. In the experiments conducted, one axis is controlled by the DRL-agent and the other by a human subject. The human subjects control their rotational axis through a tele-operation interface. A small tray, with motion capture markers attached, is given to every subject. By tilting this tray, the human operator can control one rotation axis of the robot's end-effector. A PD-controller outputs velocity commands based on the difference between the current rotation of the human's tray along the controlled axis and the rotation of the robot's tray along the same. This control loop has a delay, in addition to the inherent delay of 200ms for the DRL-agent to execute its action. This total delay complicates the task slightly for the human operator. However, forcing the human to learn the dynamics of the interaction (similarly to what the agent has to do), as opposed to normal physics as they are used to, makes for a fairer final comparison.

\subsection{Reinforcement learning system}
We implement the Soft Actor-Critic algorithm \cite{SAC1} (SAC)\footnote{Kai Arulkumaran's SAC implementation was used as foundation and modified for our purpose, Git: https://github.com/Kaixhin/spinning-up-basic}, an off-policy maximum entropy method. Running off-policy allows for the reuse of state-action transitions sampled in previous trials, which is crucial when few interaction steps are feasible. The maximum entropy framework \cite{entropy} adds an entropy maximisation term to the RL reward function, encouraging exploration. This exploration/exploitation relationship can be balanced by the temperature parameter, $\alpha$, where a larger $\alpha$ is used to encourage more exploration, and a smaller $\alpha$ corresponds to more exploitation. $\alpha$ acts as an important hyperparameter, and by using the automatic entropy tuning method introduced by Haarnoja et al. \cite{SAC}, the policy's entropy can be constrained to a desired value throughout the learning process. This removes the need for intricate hyperparameter tuning, allowing for a very sample-efficient training process. 

Using the robotic setup explained in the previous section, we define a six-dimensional state space. It consists of the position and velocity of the ball along the x and y axes in the robot's tray frame, in addition to the rotation of the robot's tray about the x and y axes. The human behaviour is included in the state space through the robot's tray rotation around the y-axis, which is mimicking the human's tray via the tele-operation interface. The action space is one-dimensional, a continuous value between -1 and 1 which is mapped to rotational velocity. For every time step $t$, the motion capture system calculates the position and orientation of both the ball and the human's tray, while the ROS transformations interface gives the robot's tray orientation. Given the observation of the current state, $s_t$, the policy network outputs a distribution of actions, from which an action, $a_t$, is sampled during training. During testing, the mean of the distribution is used, thereby removing the stochasticity, and the policy is fully exploited. The action is executed on the robot for 200ms, given that it does not move past a specific rotational limit which is defined to keep the work-space safe. The resulting state $s_{t+1}$ and reward $r_t$ is extracted, and the state transition is stored in a replay buffer of past transitions used to update the policy. A sparse reward function is used to penalise the agent with -1 for every time step and give it a reward of 10 when the target is reached. This means that reaching the target a couple of times in the early stages of training is crucial to the performance, otherwise the agent will not be able to form a representation of state values with respect to the goal. 
\subsection{Experimental setup}
Two experiments were carried out to evaluate the agent's ability to learn the interaction task. First a single subject interacted with the robot. The training process consisted of 3,500 interaction steps and 140,000 offline gradient updates using stored state transitions. The offline updates were divided throughout training, running 20,000 offline updates every 500 of real interaction steps. After every completion of such updates, the performance was tested in direct interaction with the human, averaging over ten trials. At testing, the human-agent team were given 200 interaction steps or until the target was reached. Scores were given on a scale ranging from 0 to 200: 0 means the target was not reached, 200 means the target was reached within the first time step. The entire training process, from scratch, took 45 minutes\footnote{A video of this can be seen here: \url{https://gofile.io/?c=7MoWgG}.}.

The second experiment investigated the differences in performance between human-robot and human-human collaboration with ten subjects. Every participant was allowed approximately 30 minutes of training with the robot, consisting of 3,000 interaction steps and 60,000 offline gradient updates. Ten trials of testing followed. The same human ``expert" acted as the interaction partner in all human-human trials, controlling the axis previously under control of the DRL-agent. Note that no learning was transferred from the first experiment.

\section{Results and Discussion}
Below, the learning curve from the single-subject experiment (a) and the differences in performance between the human-human and human-agent teams (b) is shown.

\begin{figure}[h]
    \centering
    \subfloat[Learning curve with standard deviation for one human-agent team.]{{\includegraphics[width=5.1cm]{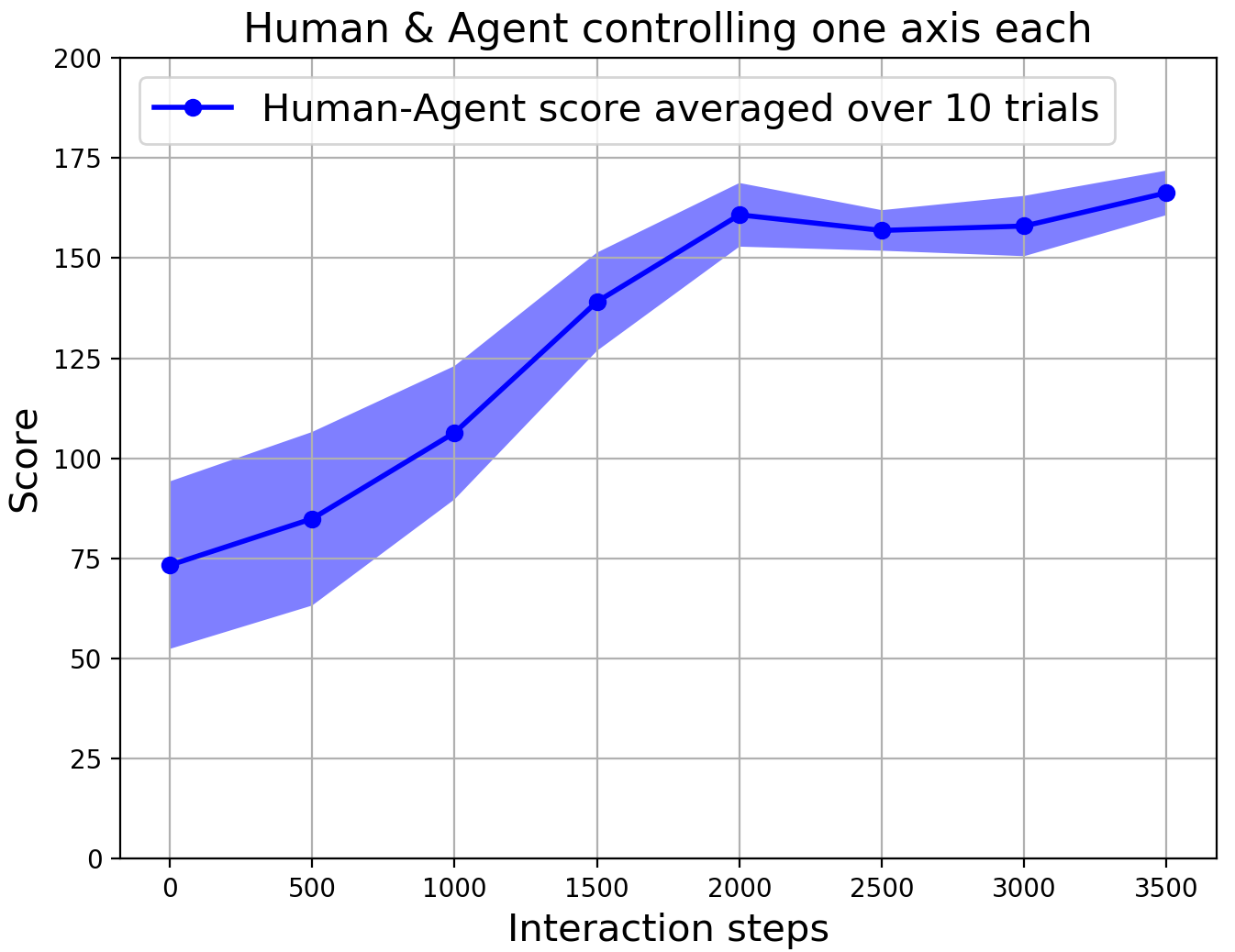} }}%
    \qquad
    \subfloat[The score and standard error of the mean for 10 subjects averaged over 10 trials.]{{\includegraphics[width=5.5cm]{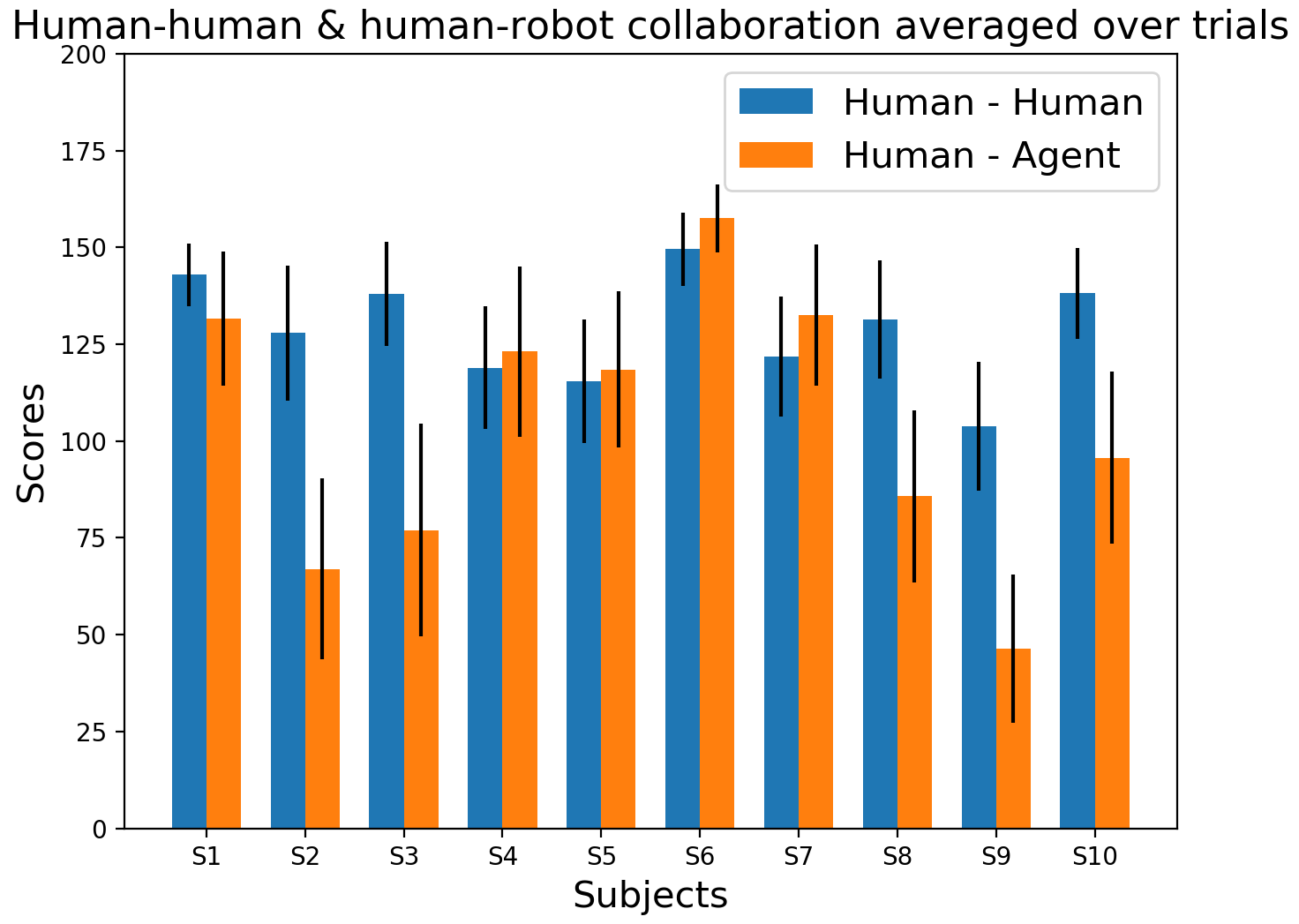} }}%
    \label{human}%
\end{figure}

Results from the single-subject experiment showed that the human-agent team was able to solve the interaction task within the time provided. Already in the first testing trial, the human-robot team were able to reach the target, even though the DRL-agent acted completely randomly. Neither of them are able to solve the task by themselves, but with human guidance, a DRL-agent can quickly build a model of the task and assist the human with only minutes of training. Furthermore, the inconsistency of the performance is decreasing as the human-robot team learn to collaborate. It is worth mentioning that the algorithm doesn't explicitly handle the fact that the human partner improves over trials, thereby changing the data the agent is trained with. However, this is equivalent to human-human collaborative learning and the fact that the agent continues to update its policy as the human partner adapts their strategy, is one of the system's strengths.

In the second experiment, the DRL-agent's ability to collaborate with humans was compared to how humans collaborate with each other. In five out of the ten subjects (S1, S4, S5, S6, S7) there was no significant differences in performance between the two scenarios. The remaining half of the subjects exhibit worse performance when collaborating with the agent. The subjects with similar performance in the two different scenarios all managed to reach the target more than twice  in the first 500 interaction steps. This was not the case for the other half of the subjects, which reached the goal at most one time. This further confirms that success in the early stages of training is vital.

\vspace{1pt}
\section{Conclusion}
We presented a robotic setup for human-robot collaborative learning implemented with a sample-efficient DRL-agent. The agent is able to solve a non-trivial collaborative task with a human partner with just 30 minutes of training and less than 4,000 interaction steps. The results proved that the agent was able to perform on a level comparable to humans, in a physical task which requires involvement from both participants to be solved. 

This work has limitations: The collaboration task is designed in such a way that the agent is not required to build personalised models to successfully complete it. Additionally, the low number of trials of random actions required to reach the target means that the same approach might be less successful in more complex tasks. Moreover, the quantitative measurement of the learning rate of the agent has an inherent limitation in its validity: it is difficult to separate between the improved performance of the agent and how the human partner learns to adapt to the system.

This work uses deep learning for the challenge of micro-data reinforcement learning, introduced by \cite{survey}.  We build human models, similar to \cite{fisac}, but with an aim of  personalised models. What sets us apart from other work is that the complete training is done with the human in the loop, from scratch, with only real-world interactions. We do not adapt a pre-trained model, as done in \cite{huang}. The results show that the traditional sample-inefficency of DRL methods does not necessarily prevent us from including humans in the training loop. A natural progression of this work would be to increase the complexity of the collaborative task, preferably including elements that require the agent to learn personalised human models, followed by extensive evaluation of these models.

\bibliographystyle{IEEEtran}
\bibliography{main.bib}

\end{document}